\documentclass[a4paper,11pt]{article}

\usepackage{fourier}
\usepackage{color}
\usepackage{graphicx}
\usepackage{url}
\usepackage[affil-it]{authblk}
\usepackage{amsmath}
\usepackage{wrapfig}

\usepackage[T1]{fontenc}
\usepackage{times}

\usepackage{subcaption}

\begin{document}

\title{FisheyeGaussianLift: BEV Feature Lifting for Surround-View Fisheye Camera Perception}


\author{Shubham Sonarghare, Prasad Deshpande, Ciaran Hogan, Deepika-Rani Kaliappan-Mahalingam, Ganesh Sistu}
\affil{Valeo Vision Systems, Tuam, Ireland}
\date{}
\maketitle
\thispagestyle{empty}

\begin{abstract}
Accurate BEV semantic segmentation from fisheye imagery remains challenging due to extreme non-linear distortion, occlusion, and depth ambiguity inherent to wide-angle projections. We present a distortion-aware BEV segmentation framework that directly processes multi-camera high-resolution fisheye images, utilizing calibrated geometric unprojection and per-pixel depth distribution estimation. Each image pixel is lifted into 3D space via Gaussian parameterization, predicting spatial means and anisotropic covariances to explicitly model geometric uncertainty. The projected 3D Gaussians are fused into a BEV representation via differentiable splatting, producing continuous, uncertainty-aware semantic maps without requiring undistortion or perspective rectification. Extensive experiments demonstrate strong segmentation performance on complex parking and urban driving scenarios, achieving IoU scores of 87.75\% for drivable regions and 57.26\% for vehicles under severe fisheye distortion and diverse environmental conditions.
\end{abstract}


\section{Introduction}
\begin{wrapfigure}{r}{0.38\textwidth}  
\vspace{-10pt}
\centering
\includegraphics[width=0.36\textwidth]{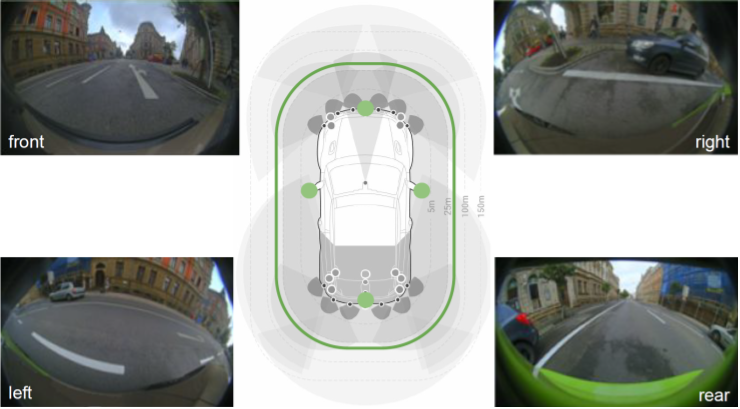}
\caption{\textit{Vehicle with 4 fisheye surround view cameras covering 360° around the vehicle \cite{yogamani2019woodscape}}}
\label{fig:camera_setup}
\vspace{-10pt}
\end{wrapfigure}

Bird’s-Eye View (BEV) representations have become fundamental to modern autonomous driving and parking systems due to their ability to spatially reason across scenes while remaining invariant to perspective distortions. BEV enables robust scene understanding by aggregating multi-view camera data into a common top-down frame, supporting tasks such as 3D object detection  \cite{huang2021bevdet}, semantic segmentation  \cite{peng2023bevsegformer}, and lane detection  \cite{chen2022persformer}.

Traditional BEV frameworks, such as Lift-Splat-Shoot (LSS)  \cite{philion2020lift}, BEVDet \cite{huang2021bevdet}, and BEVSegFormer \cite{peng2023bevsegformer}, lift image features into 3D space and project them onto a top-down grid. However, they often suffer from quantization, sparsity, or depth ambiguity—especially in distant or texture-poor regions. To mitigate these challenges, Gaussian Splatting has recently emerged as a powerful paradigm that represents scenes as ensembles of 3D Gaussians, enabling continuous and differentiable projection with learned anisotropic densities \cite{kerbl20233d}.

Building on this, GaussianLSS \cite{lu2025gaussianlss} unifies geometry-aware splatting with BEV projection by explicitly modeling depth uncertainty through Gaussian parameterization. This approach bridges photometric rendering and spatial feature pooling, offering a principled way to produce high-fidelity BEV grids from multi-camera inputs.

However, most BEV pipelines assume idealized pinhole cameras, limiting their applicability in real-world urban driving environments that often rely on wide-FOV fisheye lenses for surround coverage. Fisheye cameras offer compact, panoramic views with minimal blind spots but present strong non-linear radial distortion, complicating projection and fusion \cite{kumar2023surround}. Prior work such as FisheyeDistanceNet \cite{kumar2020fisheyedistancenet}, OmniDet \cite{kumar2021omnidet}, and more recently BEV-centric parking applications \cite{musabini2024enhanced}, demonstrate the utility of fisheye imagery in monocular depth estimation and multi-task visual perception—further motivating distortion-aware BEV perception.

Recent efforts such as 3DGUT \cite{wu20243dgut} and Fisheye-GS \cite{liao2024fisheye} have extended Gaussian Splatting to handle nonlinear imaging models, enabling accurate rendering and learning from wide angle distorted cameras. Yet, their integration into BEV perception remains underexplored. Additionally, fast and lightweight BEV inference pipelines are particularly valuable in parking environments, as highlighted by BEVFastLine \cite{narasappareddygari2024bevfastline} for single-shot BEV line detection.

In this work, we propose a novel BEV perception pipeline that leverages GaussianLSS for 3D-aware splatting and explicitly supports fisheye camera inputs, targeting high-resolution object or semantic segmentation on BEV grids. Our method addresses the dual challenge of distortion-aware projection and uncertainty-preserving spatial reasoning—key for reliable perception in dense urban and parking environments.

\section{Background}

In this section, we review relevant developments in Gaussian-based projection models and distortion-aware fisheye perception that motivate the proposed framework.

\subsection{Gaussian Scene Representations}

Gaussian-based 3D representations have recently emerged as an effective alternative to discrete voxel grids for continuous and uncertainty-aware spatial modeling. 3D Gaussian Splatting~\cite{kerbl20233d} demonstrated real-time radiance field rendering using oriented Gaussian primitives with learned anisotropic covariances, enabling differentiable rasterization and sub-pixel projection accuracy. Each 3D Gaussian is characterized by its mean $\boldsymbol{\mu} \in \mathbb{R}^3$ and covariance matrix $\boldsymbol{\Sigma} \in \mathbb{R}^{3 \times 3}$, where the corresponding spatial density function is given in equation~(\ref{eq:gaussian_density}). The covariance $\boldsymbol{\Sigma}$ allows encoding spatial uncertainty along different axes, facilitating continuous splatting into BEV grids without hard discretization.

Building upon this formulation, 3DGUT~\cite{wu20243dgut} generalized Gaussian splatting to handle general non-central camera models, including fisheye projection, by incorporating secondary ray models into the splatting operation. GaussianLSS~\cite{lu2025gaussianlss} further demonstrated how these Gaussian formulations can be employed for BEV construction by predicting per-pixel depth distributions, estimating spatial means and covariances, and aggregating projected Gaussians onto BEV grids through differentiable rasterization. Concurrently, GaussianBeV~\cite{chabot2025gaussianbev} applied Gaussian-based representations for BEV semantic segmentation directly from multi-camera perspective images, bypassing voxelization and improving BEV feature quality; however, their formulation is limited to standard pinhole cameras without addressing fisheye distortion. In contrast, our work extends these principles to distortion-aware BEV construction from wide-angle fisheye cameras for close-range driving and parking scenarios.

\subsection{Distortion-Aware Fisheye Perception}

Fisheye cameras introduce strong radial distortion effects, where the apparent scale and shape of objects vary significantly across the field-of-view. This non-uniform geometric deformation disrupts the translational equivariance assumptions commonly leveraged by standard convolutional networks, particularly for dense segmentation tasks. While undistortion-based rectification can mitigate these effects, it often leads to loss of peripheral information and resampling artifacts, especially in extreme wide-angle settings.

Several works have proposed learning-based methods for distortion-aware perception. FisheyeDistanceNet  ~\cite{kumar2020fisheyedistancenet} applied fisheye-aware monocular depth estimation, while OmniDet ~\cite{kumar2021omnidet} integrated multi-task learning for detection and segmentation under fisheye distortion. A detailed  overview of fisheye-specific challenges is provided in ~\cite{kumar2023surround}, covering calibration, scale ambiguity, and spatial resolution degradation. More recently, adaptive convolutional operators have been explored to partially compensate for spatial distortions in segmentation tasks ~\cite{manzoor2024deformable}, dynamically adjusting sampling locations within feature maps to account for local geometric variations.

To mathematically model fisheye distortion, unified projection formulations are employed. For a 3D point $\mathbf{P} = [X, Y, Z]^\top$, the distorted image coordinates $\mathbf{p} = [u, v]^\top$ are computed via:
\begin{equation}
\theta = \arctan \left( \frac{\sqrt{X^2 + Y^2}}{Z} \right)
\label{eq:theta}
\end{equation}

\begin{equation}
r = f \cdot \theta + \sum_{i=1}^{n} k_i \theta^{2i+1}
\label{eq:distortion}
\end{equation}

\begin{equation}
u = c_x + r \cdot \frac{X}{\sqrt{X^2 + Y^2}}, \quad v = c_y + r \cdot \frac{Y}{\sqrt{X^2 + Y^2}}
\label{eq:projection}
\end{equation}
Equations~(\ref{eq:theta})--(\ref{eq:projection}) describe the fisheye projection model accounting for radial distortion.

where $f$ is the focal length, $k_i$ are distortion coefficients, and $(c_x, c_y)$ is the principal point. These formulations allow accurate modeling of distorted rays for back-projection into 3D space.

Beyond purely 2D image-domain formulations, Fisheye-GS~\cite{liao2024fisheye} extended Gaussian splatting to directly accommodate fisheye camera models during 3D scene reconstruction. Nevertheless, the integration of distortion-aware Gaussian projection for dense BEV semantic segmentation from fisheye images remains an open research direction, particularly for safety-critical low speed driving applications.



\section{Methodology}

\begin{figure}[t]
  \centering
  \includegraphics[width=16cm]{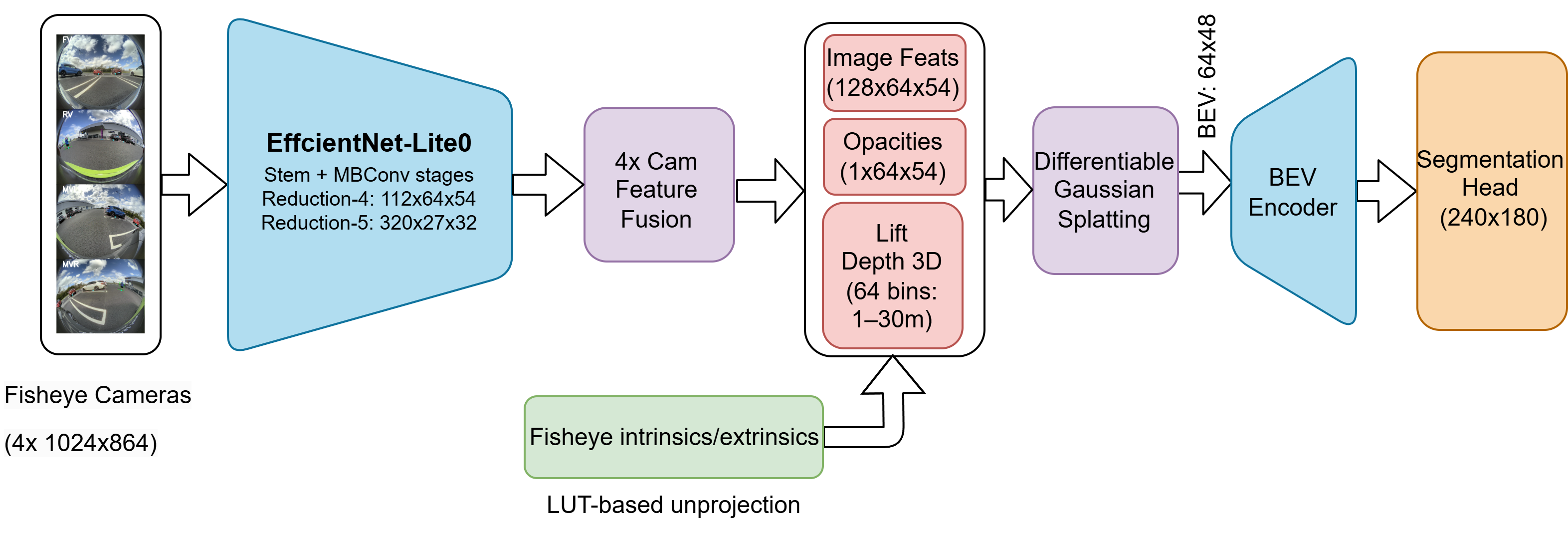}
  \caption{architecture of FisheyeGaussianLift}
  \label{fig:Fisheye_BEV_Architecture}
\end{figure}

\subsection{Model Architecture}

The model architecture for fisheye-based BEV segmentation is illustrated in Figure~\ref{fig:Fisheye_BEV_Architecture}. The architecture consists of four key stages: fisheye feature extraction, Gaussian depth projection, BEV encoding, and multi-class segmentation.

\paragraph{Fisheye Feature Extraction} 
Each fisheye image with resolution $1024 \times 864$ is processed independently through an EfficientNet-Lite backbone pretrained on ImageNet. Multi-scale features are extracted from the \texttt{reduction\_4} and \texttt{reduction\_5} blocks, yielding feature maps with channels $\{112, 320\}$. These multi-resolution features are fused using bilinear upsampling and residual bottleneck blocks, producing unified feature embeddings of dimensionality $C=128$.

\paragraph{Gaussian Depth Projection} 
For each image, depth estimation is formulated as classification over 64 uniform depth bins spanning $[1m, 30m]$. Using precomputed fisheye ray lookup tables (LUTs) derived from offline camera calibration, each image coordinate is backprojected into 3D space. The 3D coordinates $\mathbf{P}_{i,d}$ for pixel $i$ and depth bin $d$ are computed as:
\begin{equation}
\mathbf{P}_{i,d} = \mathbf{R}_i \cdot \left( \mathbf{u}_i \cdot z_d \right) + \mathbf{t}_i,
\end{equation}
where $\mathbf{u}_i$ is the unit direction from the LUT, $z_d$ is the center of the $d$-th depth bin, and $(\mathbf{R}_i, \mathbf{t}_i)$ are the extrinsic parameters of the fisheye camera.

Each projected 3D point is then parameterized as a Gaussian distribution \cite{kerbl20233d} $\mathcal{G}(\mathbf{x})$ with mean $\boldsymbol{\mu} = \mathbf{P}_{i,d}$ and covariance $\boldsymbol{\Sigma}$ defined as:
\begin{equation}
\mathcal{G}(\mathbf{x}) = \frac{1}{(2\pi)^{3/2} |\boldsymbol{\Sigma}|^{1/2}} 
\exp \left( -\frac{1}{2} (\mathbf{x} - \boldsymbol{\mu})^\top \boldsymbol{\Sigma}^{-1} (\mathbf{x} - \boldsymbol{\mu}) \right).
\label{eq:gaussian_density}
\end{equation}

The covariance $\boldsymbol{\Sigma}$ models both learned depth uncertainty and discretization-induced variance \cite{lu2025gaussianlss, wu20243dgut}:
\begin{equation}
\boldsymbol{\Sigma} = \sigma^2 \mathbf{I}_3 + \Delta_d \Delta_d^\top,
\end{equation}
where $\sigma$ is a network-predicted uncertainty scalar, and $\Delta_d$ reflects the discretization error from depth bin quantization.

The resulting 3D Gaussians are then splatted into BEV space using anisotropic differentiable rasterization. \cite{kerbl20233d}

\paragraph{BEV Encoding}
The splatted Gaussian features from multiple fisheye views are aggregated into a unified top-down BEV grid covering $24\,m \times 18\,m$, discretized into a $64 \times 48$ grid with $0.375\,m$ resolution. The BEV encoder is composed of three ResNet blocks same as LSS\cite{philion2020lift}. Each downsampling block applies strided convolutions followed by bottleneck residual layers, while upsampling uses bilinear interpolation fused with convolutional layers to recover spatial resolution and aggregate multi-scale spatial context.

\paragraph{Multi-Class Segmentation Head}
The final BEV feature map is passed to a semantic segmentation head that predicts dense semantic maps including drivable areas and vehicles, which form the primary evaluation classes. The segmentation head employs transposed convolutions to progressively upsample the feature map to the final output resolution of $240 \times 180$. A weighted cross-entropy loss is applied to address class imbalance across categories. The entire network is trained end-to-end with supervision on BEV semantic labels.

\section{Dataset}
\label{sec:dataset}

The dataset consists of multi-fisheye surround-view images collected in real-world parking lots, private driving areas, and low-speed urban environments. Our vehicle sensor setup is similar to \cite{yogamani2019woodscape} and consists of a four-camera fisheye system that captures synchronized images from Front View (FV), Rear View (RV), Mirror-View Right (MVR) and Mirror-View Left (MVL), each with a resolution of $1024 \times 864$ pixels and an approximate field of view of $190^\circ$. All images are timestamp-synchronized with high-accuracy vehicle odometry for precise spatial referencing during annotation.

The dataset contains approximately $4790$ distinct driving traces, partitioned into $3832$ sequences for training, $479$ for validation, and $479$ for evaluation. Each sequence consists of continuous timesteps that capture full 360-degree fisheye surround-view imagery during parking maneuvers and low-speed urban driving, allowing spatial-temporal consistency in diverse scenarios. In total, the training and evaluation sets contain $59675$ and $9000$ timesteps, corresponding to $238k$ and $72k$ total images on all camera views.

Data were collected across multiple geographic regions, including the Czech Republic (CZE, 416 traces), Germany (DEU, 1014), Ireland (IRL, 802) and the United States (USA, 2558), ensuring diverse scene layouts, road markings, signage, and environmental structures. The recordings span a variety of parking layouts, such as city lots, indoor garages, open outdoor parking, urban streets, and rural driveways. To ensure system generalizability, data were collected using six different vehicle platforms with varying camera rigs.

The data set comprehensively covers the environmental variations critical for parking scenarios, including illumination conditions: artificial indoor lighting, low-light dawn / dark, dark unlit regions, street light scenarios  and bright daylight. Weather diversity includes dry, light rain, heavy rain, light snow, and fog/haze sequences. The surface diversity of the scene includes asphalt, cobbled, paving stones, gravel, grass, reflective surfaces, and varying wet/dry conditions.

The camera intrinsic parameters, fisheye distortion coefficients and extrinsics were calibrated offline using a multiplane checkerboard method, producing complete fisheye projection models used for both annotation and projection during training. Pixel-level manual semantic annotations were performed in the fisheye image space and later mapped to BEV representations for supervised training.

\section{Experimental Setup}
\label{sec:training}

All experiments are implemented in PyTorch using a single Quadro RTX 8000 (50G). The input fisheye images are processed at their native resolution of $1024 \times 864$ without distortion correction.

The EfficientNet-L0 encoder is initialized with pretrained ImageNet weights. Training is performed in two stages: (1) the neck, Gaussian projection, and segmentation head are trained for 30 epochs while freezing the backbone, followed by (2) joint end-to-end fine-tuning with the backbone unfrozen for 20 additional epochs. A batch size of 4 is used, with each epoch requiring approximately 3 hours of training time.

The AdamW optimizer is employed with an initial learning rate of $3 \times 10^{-4}$ and weight decay of $1 \times 10^{-7}$. A one cosine cycle learning rate scheduler modulates the learning rate throughout training. To improve generalization, data augmentation includes random brightness, contrast, saturation adjustments, Gaussian noise injection, horizontal flipping, scaling, and minor rotations.

\subsection{Loss Function}

The network is trained using a weighted cross-entropy loss applied to the BEV semantic segmentation output. The loss for each pixel $i$ and class $c$ is defined as:
\begin{equation}
\mathcal{L}_{\text{seg}} = - \sum_{i} \sum_{c} w_c \, y_{i,c} \log p_{i,c},
\label{eq:loss}
\end{equation}
where $y_{i,c}$ denotes the multi-label mask for the class $c$ at grid cell $i$, $p_{i,c}$ is the predicted softmax probability, and $w_c$ is the class-specific weight to address class imbalance.

No auxiliary losses such as center heatmap regression or offset regression are employed. The network is optimized end-to-end using this weighted cross-entropy objective to directly supervise dense multi-class BEV segmentation.

\subsection{Evaluation Metrics}

Our model performance is evaluated using class-wise Intersection-over-Union (IoU), computed for each semantic class $c$ directly on the BEV outputs. The IoU is defined as:
\begin{equation}
\text{IoU}_c = \frac{TP_c}{TP_c + FP_c + FN_c},
\end{equation}
where $TP_c$, $FP_c$, and $FN_c$ denote the true positives, false positives, and false negatives for class $c$, respectively.

The evaluation focuses on critical semantic classes including drivable regions and vehicles. All metrics are computed on the predicted BEV semantic maps at full output resolution ($240 \times 180$ grid cells), corresponding to a $24\,\mathrm{m} \times 18\,\mathrm{m}$ spatial coverage. Reported scores are averaged over the entire test dataset.

\section{Experimental Results}

In this section, we present the experimental evaluation of the proposed fisheye based BEV semantic segmentation model using Gaussian projection. The evaluation is carried on the described private parking dataset acquired using a surround-view fisheye camera system, following the experimental protocol detailed in Section~\ref{sec:dataset} and Section~\ref{sec:training}.

\subsection{Quantitative Results}

Table~\ref{tab:results} summarizes the BEV segmentation performance of the proposed approach alongside several existing methods from the literature. It should be noted that the listed methods operate on different datasets, camera models, and image resolutions; thus, the results are not strictly comparable in absolute terms. Nevertheless, the table is provided to offer a relative perspective on achievable performance levels across different system configurations.

The proposed fisheye-based architecture achieves \textbf{87.75}\% IoU for drivable region segmentation and \textbf{57.26}\% IoU for vehicle segmentation. These results are obtained on a proprietary fisheye dataset using high-resolution inputs ($1024 \times 864$), compared to prior works such as GaussianLSS~\cite{lu2025gaussianlss} and CVT~\cite{chen2022persformer} that operate on perspective cameras from the nuScenes or Lyft datasets at lower resolutions ($448 \times 800$). Despite these differences, the superior performance highlights the advantages of jointly modeling fisheye distortion, depth uncertainty, and continuous Gaussian splatting.

The combination of LUT-based distortion-aware unprojection, probabilistic Gaussian depth modeling, and differentiable splatting \cite{kerbl20233d} enables robust BEV feature alignment even under severe lens distortion. The resulting high-fidelity BEV grids contribute to improved segmentation consistency across complex parking layouts, dense scenes, and diverse environmental conditions.

\begin{table}[!h]
\begin{center}
\footnotesize  
\begin{tabular}{|c|c|c|c|c|c|c|}
\hline
Method & Backbone & Dataset & Resolution & Camera & Drivable IoU (\%↑) & Vehicle IoU (\%↑) \\
\hline
LSS [Philion and Fidler, 2020] & EN-B0 & Lyft & 448$\times$800 & Pinhole & 72.94 & 44.64 \\
CVT [Chen et al., 2022] & EN-B4 & nuScenes & 448$\times$800 & Pinhole & 74.3 & 36.1 \\
GaussianLSS [Lu et al., 2025] & EN-B4 & nuScenes & 448$\times$800 & Pinhole & 76.3 & 40.6 \\
\textbf{Ours} & EN-L0 & Proprietary & 1024$\times$864 & Fisheye & \textbf{87.75} & \textbf{57.26} \\
\hline
\end{tabular}
\end{center}
\vspace{-8pt}
\caption{BEV segmentation performance comparison across prominent published architectures, datasets, and camera modalities.}
\label{tab:results}
\vspace{-10pt}
\end{table}

\subsection{Qualitative Analysis}

\begin{figure}[t]
\centering
\begin{subfigure}{0.48\linewidth}
  \includegraphics[width=\linewidth]{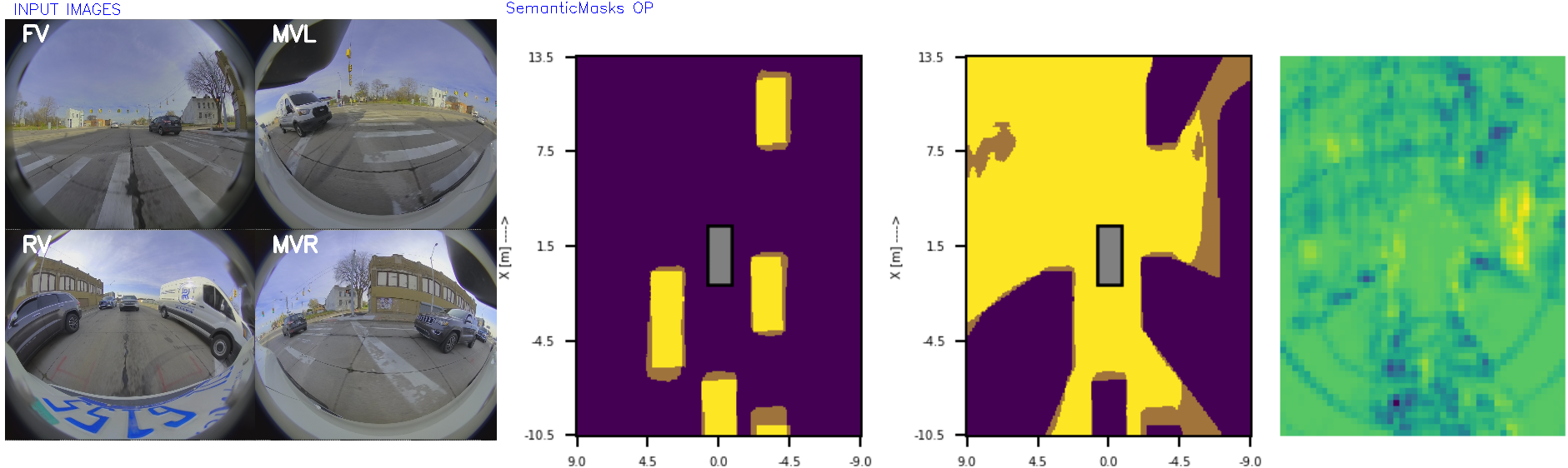}
\end{subfigure}\hfill
\begin{subfigure}{0.48\linewidth}
  \includegraphics[width=\linewidth]{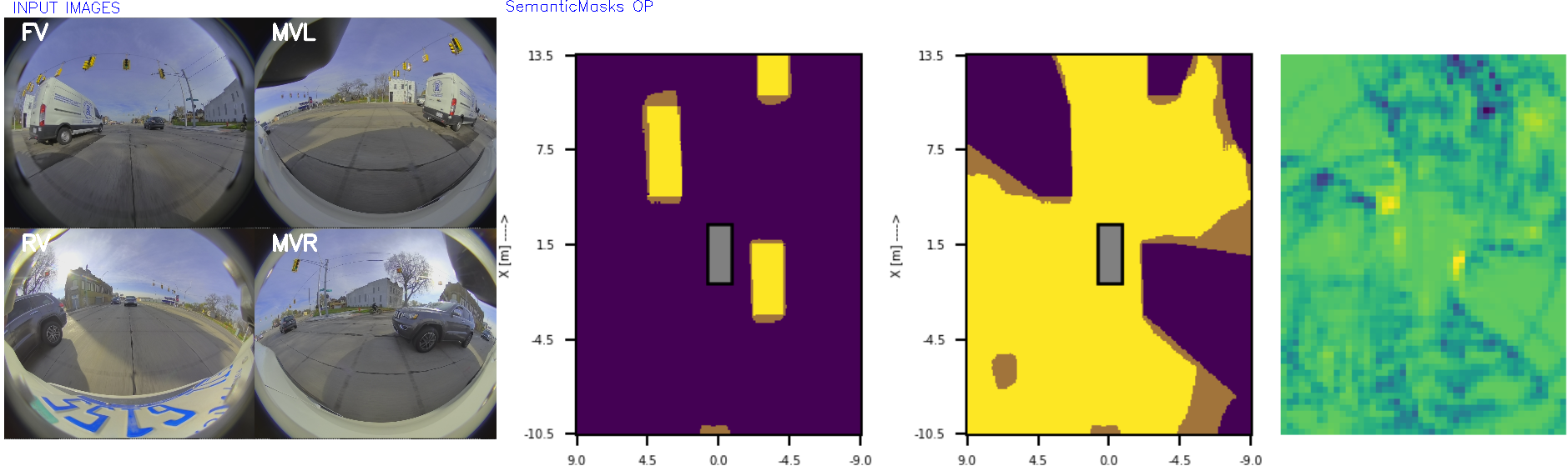}
\end{subfigure}
\vspace{3pt}
\begin{subfigure}{0.48\linewidth}
  \includegraphics[width=\linewidth]{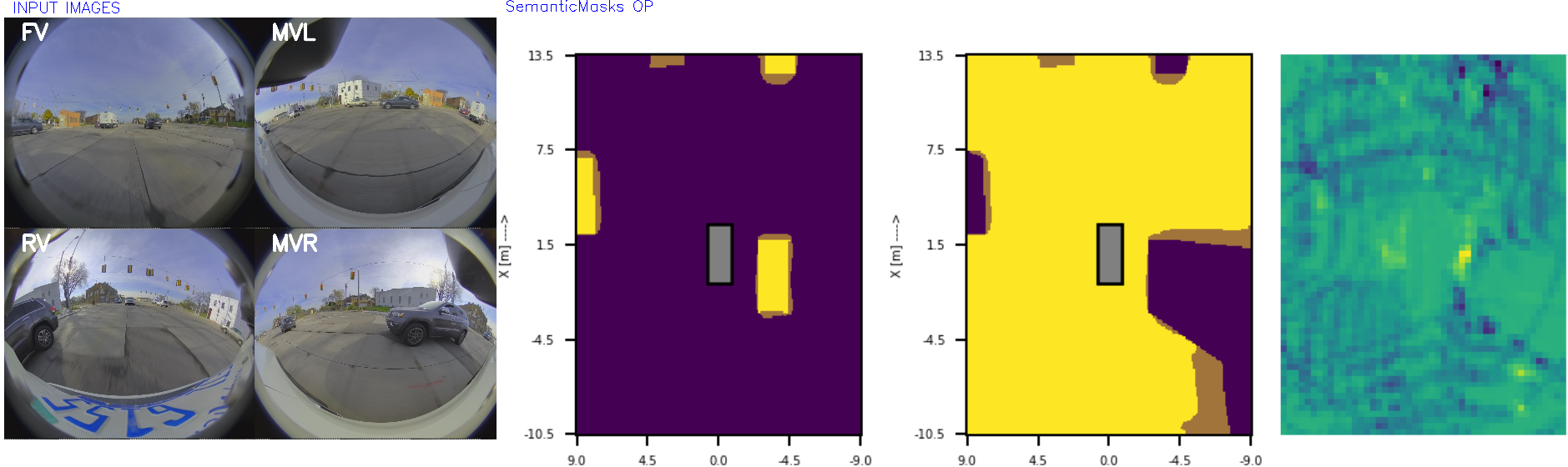}
\end{subfigure}\hfill
\begin{subfigure}{0.48\linewidth}
  \includegraphics[width=\linewidth]{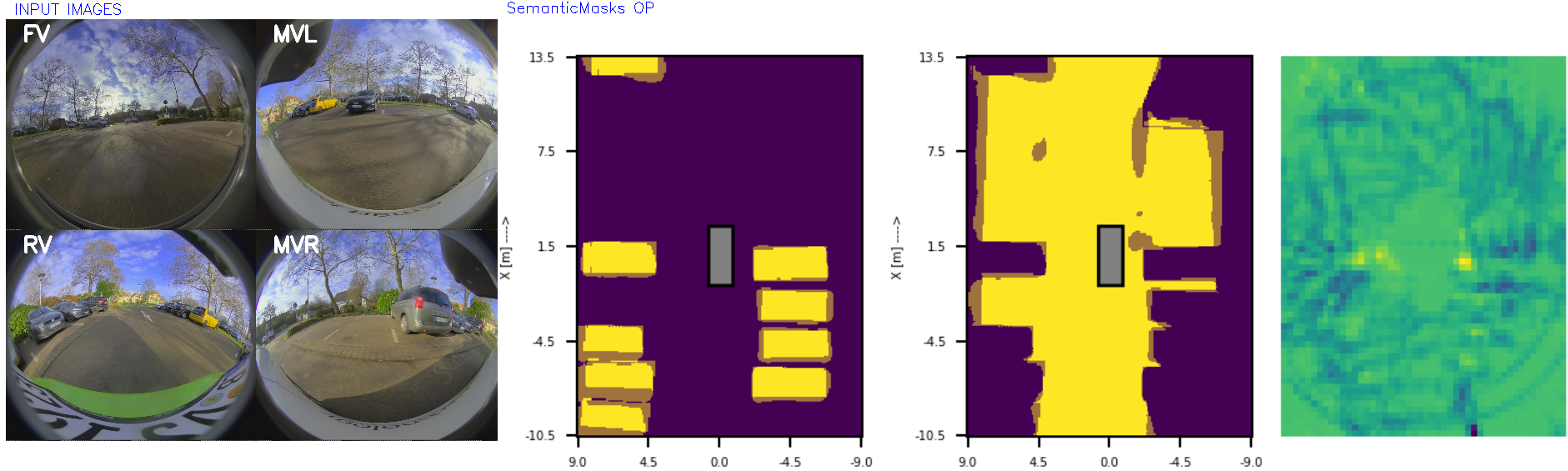}
\end{subfigure}
\vspace{-5pt}
\caption{Qualitative BEV segmentation results on fisheye images. Each visualization shows 4 fisheye camera views (FV, RV, MVL, MVR), predicted BEV segmentation masks for vehicle and drivable space classes, and the intermediate BEV feature activation map.}
\label{fig:qualitative_results}
\vspace{-10pt}
\end{figure}

Figure~\ref{fig:qualitative_results} presents representative qualitative results obtained from the validation set. Each example illustrates the fisheye surround-view inputs (FV, RV, MVL, MVR) at $1024 \times 864$ resolution, followed by corresponding BEV segmentation outputs and feature activations. 

The qualitative examples demonstrate that the model accurately localizes parked vehicles and delineates drivable regions across varying parking layouts and urban intersections. Despite the strong fisheye distortions, the proposed Gaussian projection framework successfully preserves spatial consistency in object boundaries after unprojection into BEV space. The BEV feature maps exhibit distinct activations around object regions, indicating effective multi-camera feature fusion and geometric reasoning.

The model exhibits robustness across diverse illumination and scene contexts, including varying sun positions, shadows, textured pavements, and surrounding urban structures. Importantly, the segmentation outputs remain geometrically consistent across overlapping fields-of-view and vehicle orientations, reflecting the reliability of the fisheye-specific unprojection and differentiable Gaussian splatting pipeline.

\section{Conclusion}

Our work FisheyeGaussianLift demonstrates effective handling of wide-angle fisheye distortion through direct incorporation of precomputed LUT-based unprojection, ensuring accurate 3D ray backprojection without requiring explicit image rectification. The depth discretization strategy combined with Gaussian uncertainty modeling enables stable spatial lifting, while differentiable Gaussian splatting provides smooth feature aggregation into BEV space without voxelization artifacts. This continuous formulation supports sub-grid alignment of spatial features, which is particularly critical in high-resolution parking scenarios.

The efficient EfficientNet-L0 backbone, combined with feature fusion and Gaussian projection, enables accurate BEV semantic segmentation while maintaining computational efficiency. The end-to-end design is highly modular, allowing for straightforward extension to additional sensor modalities and larger camera configurations. The results demonstrate that distortion-aware Gaussian projection offers a robust mechanism for precise BEV map generation directly from multi-camera fisheye imagery, addressing key challenges in dense urban parking environments.

\section*{Acknowledgments}

The authors thank Valeo Vision System, Tuam, Ireland for their support, resources, and the opportunity to contribute tothe broader research community with this work


\bibliographystyle{apalike}

\bibliography{imvip}

\end{document}